\documentclass[journal]{IEEEtran}
\usepackage{cite}

\ifCLASSINFOpdf
 \usepackage[pdftex]{graphicx}
\else
\fi
\usepackage{amsmath}
\usepackage{multirow}

\usepackage{algorithmic}

\usepackage{array}

\ifCLASSOPTIONcompsoc
 \usepackage[caption=false,font=normalsize,labelfont=sf,textfont=sf]{subfig}
\else
 \usepackage[caption=false,font=footnotesize]{subfig}
\fi
\usepackage{fixltx2e}

\usepackage{stfloats}
\ifCLASSOPTIONcaptionsoff
 \usepackage[nomarkers]{endfloat}
\let\MYoriglatexcaption\caption
\renewcommand{\caption}[2][\relax]{\MYoriglatexcaption[#2]{#2}}
\fi
\usepackage{url}

\usepackage{enumitem}

\newtheorem{theorem}{Theorem}

\begin{document}




%
\title{Learning from Matured Dumb Teacher\\for Fine Generalization}
%
%
%

\author{HeeSeung~Jung,~
        Kangil~Kim,~\IEEEmembership{Member,~IEEE,}~Hoyong~Kim,~and~Jong-Hun~Shin
\thanks{HeeSeung Jung, Kangil Kim, and Hoyong Kim are with the School of Electrical Engineering and Computer Science, Gwangju Institute of Science and Technology, Gwangju, South Korea (corresponding author: kangil.kim.01@gmail.com). Jong-Hun Shin with Electronics and Telecommunications Research Institute, Yuseong-Gu, Gajeong-Ro 218, Deajun, 34129, South Korea.}
}

%
%


\markboth{Submitted to IEEE TRANSACTIONS ON NEURAL NETWORKS AND LEARNING SYSTEMS}%
{Jung \MakeLowercase{\textit{et al.}}: Learning from Matured Dumb Teacherfor Fine Generalization}

%



\maketitle

\begin{abstract}
The flexibility of decision boundaries in neural networks that are unguided by training data is a well-known problem typically resolved with generalization methods. 
A surprising result from recent knowledge distillation~(KD) literature is that random, untrained, and equally structured teacher networks can also vastly improve generalization performance. 
It raises the possibility of existence of undiscovered assumptions useful for generalization on an uncertain region.
In this paper, we shed light on the assumptions by analyzing decision boundaries and confidence distributions of both simple and KD-based generalization methods.
Assuming that a decision boundary exists to represent the most general tendency of distinction on an input sample space (i.e., \textit{the simplest hypothesis}), we show the various limitations of methods when using the hypothesis.
To resolve these limitations, we propose \textit{matured dumb teacher based KD}, conservatively transferring the hypothesis for generalization of the student without massive destruction of trained information.
In practical experiments on feed-forward and convolution neural networks for image classification tasks on MNIST, CIFAR-10, and CIFAR-100 datasets, the proposed method shows stable improvement to the best test performance in the grid search of hyperparameters. 
The analysis and results imply that the proposed method can provide finer generalization than existing methods.
\end{abstract}


%
\IEEEpeerreviewmaketitle

\section{Introduction}
%
%
%
%
\IEEEPARstart{G}{eneralization} methods for neural networks entail $\ell_1$ or $\ell_2$ penalization, dropout, and label smoothing~\cite{srivastava2014dropout,szegedy2016rethinking}, usually allowing flexible decision boundaries, even when using the same strength of generalization and training loss. 
Under these conditions, a boundary is arbitrarily selected, and obtaining a better generalized model is not guaranteed. 

Self-knowledge distillation~(sKD)~\cite{kim2020self, hahn2019self, zhang2019your, yun2020regularizing,furlanello2018born} is a potential data-driven generalization method to resolve the problem. 
sKD follows the basic transferring mechanism of knowledge distillation (KD)~\cite{hinton2015distilling, muller2019does}, but it builds a teacher model, whose architecture is equal to the student, without the help of external resources. 
The sKD mechanism used to transfer knowledge from a separately trained good teacher to a student is similar to ensemble mechanisms, such that the generalization effect is naturally induced. 
Surprisingly, even poor teachers can effectively generalize student models in sKD~\cite{yuan2020revisiting,pereyra2017regularizing}.
However, the dynamics required for understanding the causes and effects of such teachers as a solely working generalization method are still in a veil of mystery. 
This issue raises an old question, ''what is the most probable assumption that can be constructed from data to generalize on the uncertain region that otherwise allows the flexibility of decision boundaries?''

In this paper, we shed light on this question by analyzing existing generalization methods using \textit{the simplest hypothesis}, which is a hypothetical decision boundary built by a strongly generalized model on a low-quality local optimum.
It becomes a probable hypothesis when there is no better assumption on the uncertain area other than random selection. 
From the perspective of adopting the hypothesis, well-known simple distillation-based generalization methods have limitations, including a bias on decision boundary updates or the blurring of confidence distributions.
We analyze these limitations and propose a new way of using the simplest hypothesis: \textit{matured dumb Teacher based Knowledge Distillation} (mDT-KD). 
The key idea is to extract a hypothesis from the matured dumb teacher (mDT) and conservatively transfer it to the student.
This method induces \textit{fine} generalization to preserve an accurate confidence distribution constructed from training data and to adapt the simplest hypothesis to the uncertain area. 
We empirically analyze the differences of generalization methods of a simple binary classification task suitable for directly analyzing decision boundaries and confidence distributions. 
The achievable performance is investigated with an intensive grid search for more practical image classification tasks on the MNIST \cite{lecun-mnisthandwrittendigit-2010}, CIFAR-10, and CIFAR-100 \cite{krizhevsky2009learning} datasets. 
In summary, our contributions are as follows: 
\begin{itemize}[noitemsep,topsep=0pt,parsep=0pt,partopsep=0pt]
\item the simplest hypothesis is proposed as the most probable assumption about the uncertain region to determine the decision boundaries;
\item the limitations of both simple and distillation-based generalization methods are clarified for using the simplest hypothesis;
\item mDT-KD is proposed to effectively build and use the simplest hypothesis;
\item the generalization impact of decision boundaries and confidence distributions are empirically analyzed for an easy-to-understand problem;
\item practical performance in well-known image classification benchmarks is evaluated using an intensive grid-search algorithm with hyperparameters that affect generalization strength.

\end{itemize}

\section{Related Work} \label{sec:relatedwork}
\subsection{Knowledge Distillation}
Knowledge Distillation (KD) is a method of transferring information from a larger and cumbersome teacher model to a new student model, which induces good performance even that student is a small and shallow model~\cite{romero2014fitnets, heo2019knowledge, mirzadeh2020improved}.
The main factor of this effect is to transfer dark knowledge of the teacher as a smoothed soft label.~\cite{guo2017calibration, hinton2015distilling}
To effectively use the dark knowledge, the softmax function is modified as follows:
\begin{equation}
\label{eq:ssoftmax}
softmax(x_i, T) = \frac{e^{x_i/T}}{\Sigma_j^N e^{x_j/T}},
\end{equation}
where $T$ is temperature, and $N$ is the number of classes.
The new loss function that uses both hard and soft labels is defined as 
\begin{equation}
\label{eq:naive_kd}
\mathcal{L}_{KD} = \mathcal{L}_{CE}(\sigma(y_s), y_{gt})\\
+ \lambda \cdot \mathcal{L}_{CE}(\sigma(\frac{y_s}{T}), \sigma(\frac{y_t}{T})),
\end{equation}
where $\mathcal{L}_{CE}$ is the cross-entropy loss, $y_t$ and $y_s$ are the output probability of the teacher and student, and $y_{gt}$ is the ground truth.

\subsection{Self-Knowledge Distillation}
Self-knowledge distillation~(sKD) uses the same architecture as KD for the teacher and student, but it is more oriented to generalization rather than model compression.
In~\cite{kim2020self, furlanello2018born}, an sKD was repeated between suboptimal models observed in a training pass, inducing ensemble-like effects. 
This method also distills confidence distributions over classes of different samples having the same labels during training~\cite{yun2020regularizing}, which is typical of cases using dark knowledge for generalization. 
Beyond the practical uses, we focus on the reasons of good generalization when using sKDs while further proposing which teacher models to use.

\subsection{Label Smoothing}
Label smoothing~(LS)~\cite{szegedy2016rethinking} generalizes a model by weakening confidence on correct labels.
The method smooths the confidence of a hard-label in one-hot representation by interpolating it with a uniform distribution, redefined as:
\begin{equation}
\label{eq:ls}
y_{ls} = (1 - \epsilon) \cdot y_{gt} + \epsilon \cdot \frac{\vec{1}}{N}, 
\end{equation}
where $\epsilon$ is the hyperparameter of this method, and $N$ is the number of classes.
LS can retain the correct label as the most probable one, although it adds uncertainty. 
The method results in performance improvement in some classification tasks~\cite{pereyra2017regularizing}, but it may be limited in detailed conditions as analyzed in~\cite{muller2019does}.
We regard LS as another type of KD in this paper. 

\section{Difficulty of Learning the Simplest Hypothesis in Generalization Methods} \label{sec:limit_of_reg}

In the terminology of this paper, the \textit{hypothesis} indicates the set of hyperplanes needed to build a decision boundary on an input space for different classes.
\textit{The simplest hypothesis} implies a trained decision boundary of a low-quality local optimal model obtained through strong generalization. 
In this section, we raise the difficulties of existing generalization methods in two groups: those solely working on simple methods and sKD-based ones using a teacher model. 

\subsection{Simple Generalization}
The widely used $\ell_1$ and $\ell_2$ penalization, dropout, and LS come with different types of difficulties regarding the simplest hypothesis. 
\subsubsection{Penalization}

\begin{figure} 
\centering
\subfloat[bias of penalization\label{fig:penalization_geometric_a}]{%
 \includegraphics[width=4.5cm]{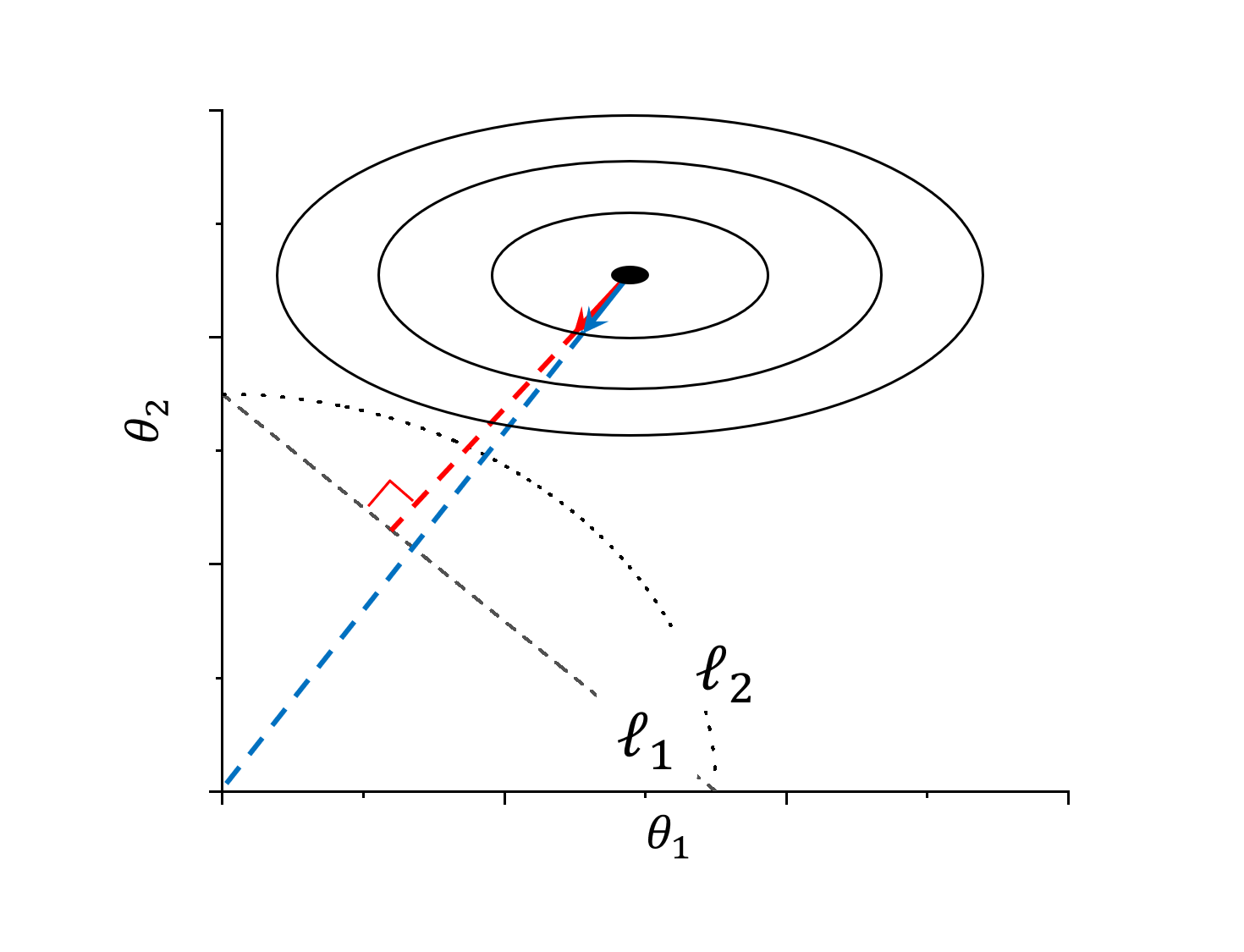}}
\subfloat[teacher directions in sKD methods\label{fig:penalization_geometric_b}]{%
\includegraphics[width=4.5cm]{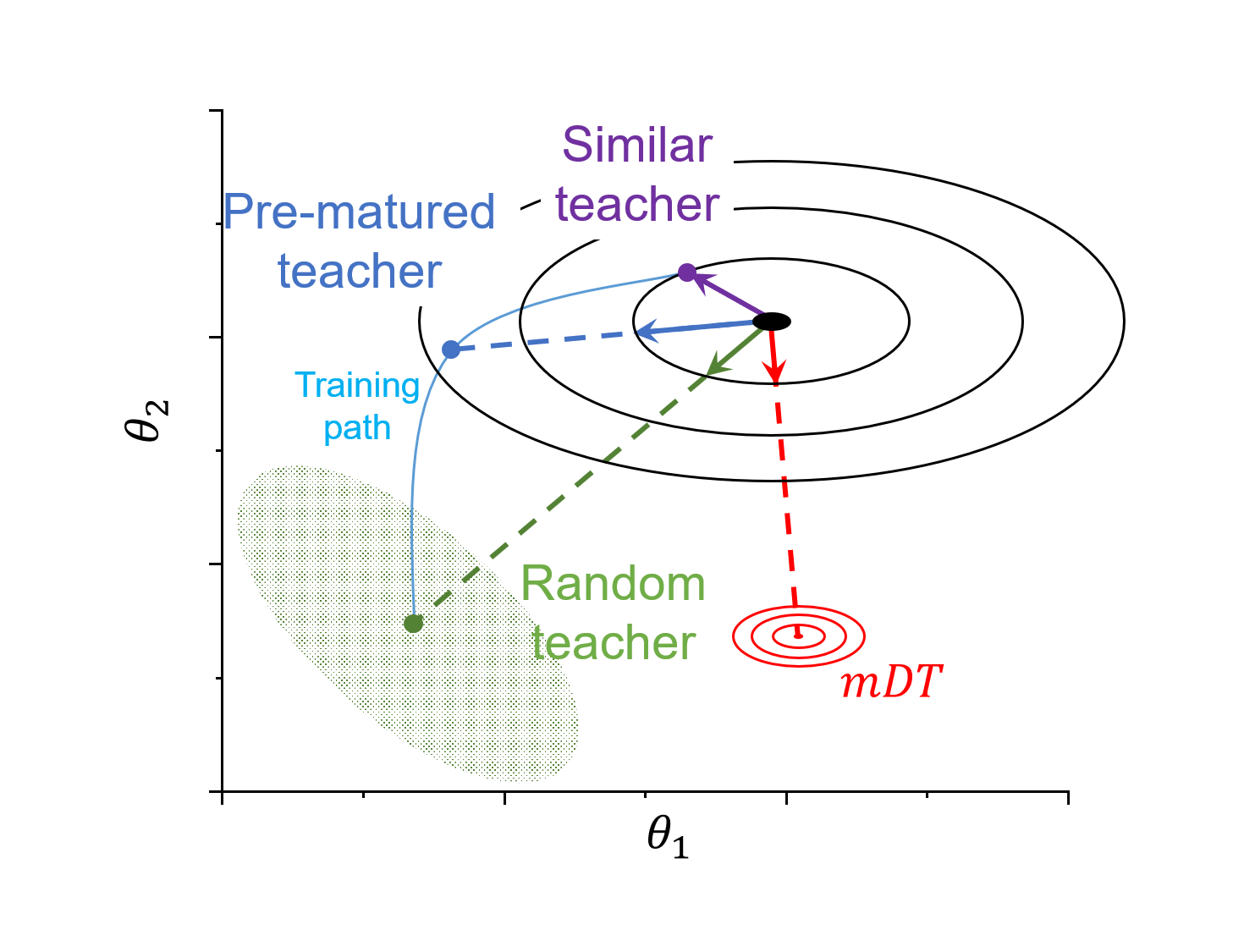}}
\caption{Update directions in the uncertain region on the training loss landscape. The black point is the trained student model without generalization. The flat optima is the uncertain region of generalization. (a) shows the data-agnostic bias of $\ell_1$ and $\ell_2$ penalization, and (b) shows an example of the different directions of teacher models of the sKD methods. ($\theta_i$: model parameters).}
\label{fig:penalization_geometric} 
\end{figure}

Penalization is a method of adding a \textit{penalty} regarded as model complexity. 
The simple and generally used penalties are $\ell_1$-norm based penalty $\lambda\sum\limits |w_i|$ and $\ell_2$-norm based penalty $\frac{\lambda}{2} \sum\limits w_i^2$, also known as Lasso and Ridge regularization or weight decaying.
Their scales are empirically tuned by $\lambda$. 
The two difficulties of this approach involve (1) biases of changing decision boundaries and (2) the difficulty of using the information of a low-quality optimum in a single model. 

The main difficulty is the bias of gradients irrelevant to the simplest hypothesis.
Figure~\ref{fig:penalization_geometric}(a) shows the landscape of a training loss and penalties over the space of a model $\theta= \theta_1 \cup \theta_2$. 
Let $J(\theta)$ be the training loss and $M_x$ be sets of assigned values for $\theta$.
If there exists a path from $M_i$ such that $J(M_i) = J(M_j)$ for all $M_j$ on the path, then the covered area by all possible paths is defined as a \textit{plateau}.
\begin{theorem}
In any plateau of a good local optimum, $M_{h}$, $\ell_1$ and $\ell_2$ penalization do not guarantee to move toward a worse but simple local optimum $M_{l}$ which indicates the simplest hypothesis. 
\IEEEproof 
$\nabla_\theta\ell_1=\vec{1}$ and
$\nabla_\theta\ell_2=\theta$ for all $M$. $M_l$ varies by training data and its cost, but the two gradients are independent of the data and cost function. 
\end{theorem}
The bigger problem is that penalization can not inform where $M_l$ exists unless we break $M_h$.
\begin{figure}[htb]
\center{\includegraphics[width=0.5\textwidth]{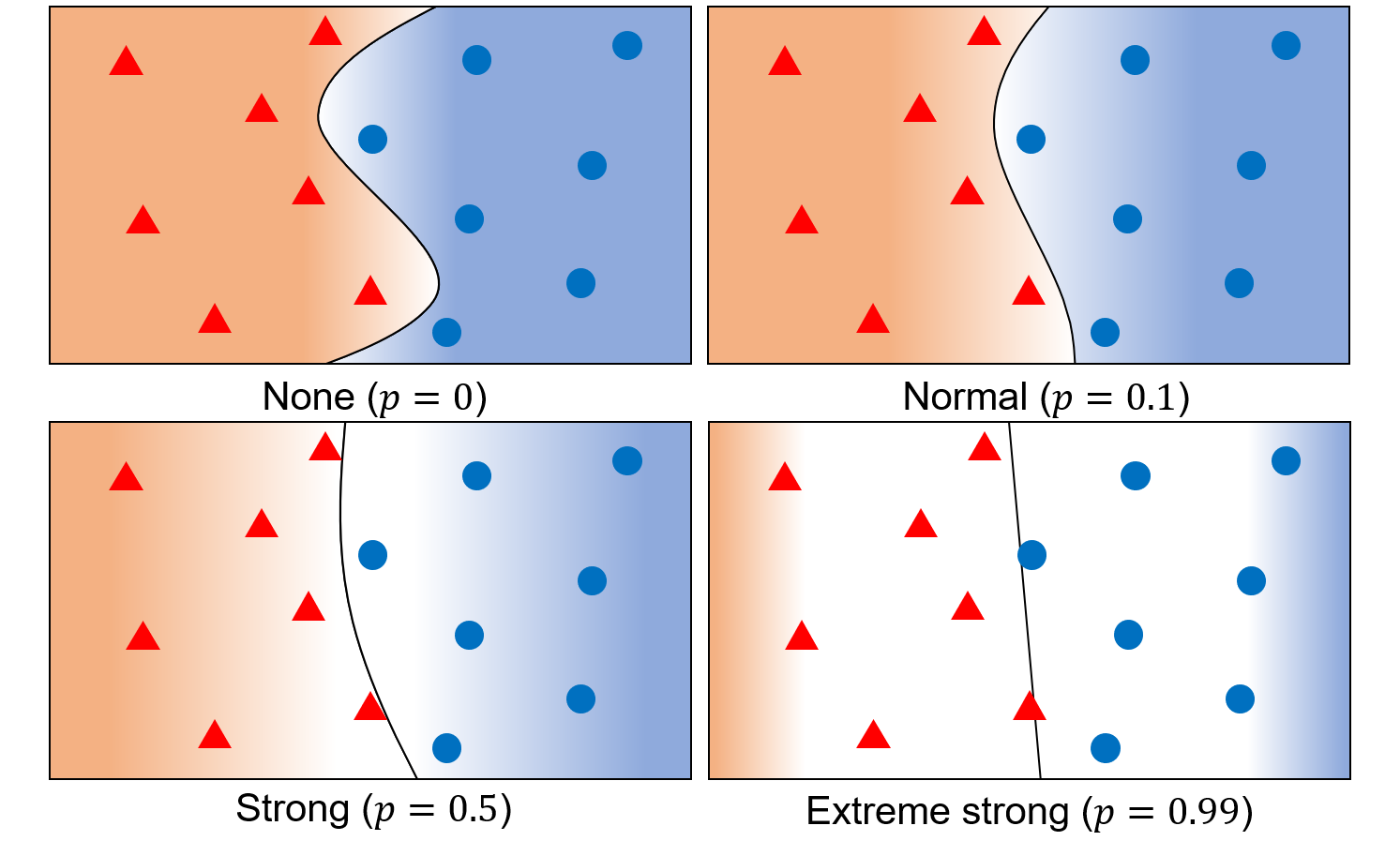}}
\caption{Decision boundary and confidence distribution of dropout of various strength. $p$ is the dropout probability to turn off a node. Red and black background colors are predictions for each class. White tubes represent confidence. The lighter color is less confident. }
\label{fig:dropout_theory}
\end{figure}

\subsubsection{Dropout}
In the case of dropout, the simplest hypothesis can be smoothly constructed by adjusting the probability, $p$, of turning off a node, but the confidence is largely sacrificed~\cite{zhang2019confidence}.
Using dropout can be interpreted as an ensemble model comprising a large number of small networks~\cite{srivastava2014dropout}. 
From another perspective, a single component of the ensemble is responsible for more parts of decision boundaries in a larger $p$. 
Hence, the small networks build simpler hyperplanes because of their finite capacity. In Figure~\ref{fig:dropout_theory}, the decision boundary flattening is shown while $p$ increases. 
Simplifying the decision boundary can be regarded as the simplest hypothesis, but it is obtained by sacrificing confidence on the prediction, as shown by the broadened white-colored tube enwrapping the decision boundary.
\subsubsection{LS}
\begin{figure}[htb]
\center{\includegraphics[width=0.45\textwidth]{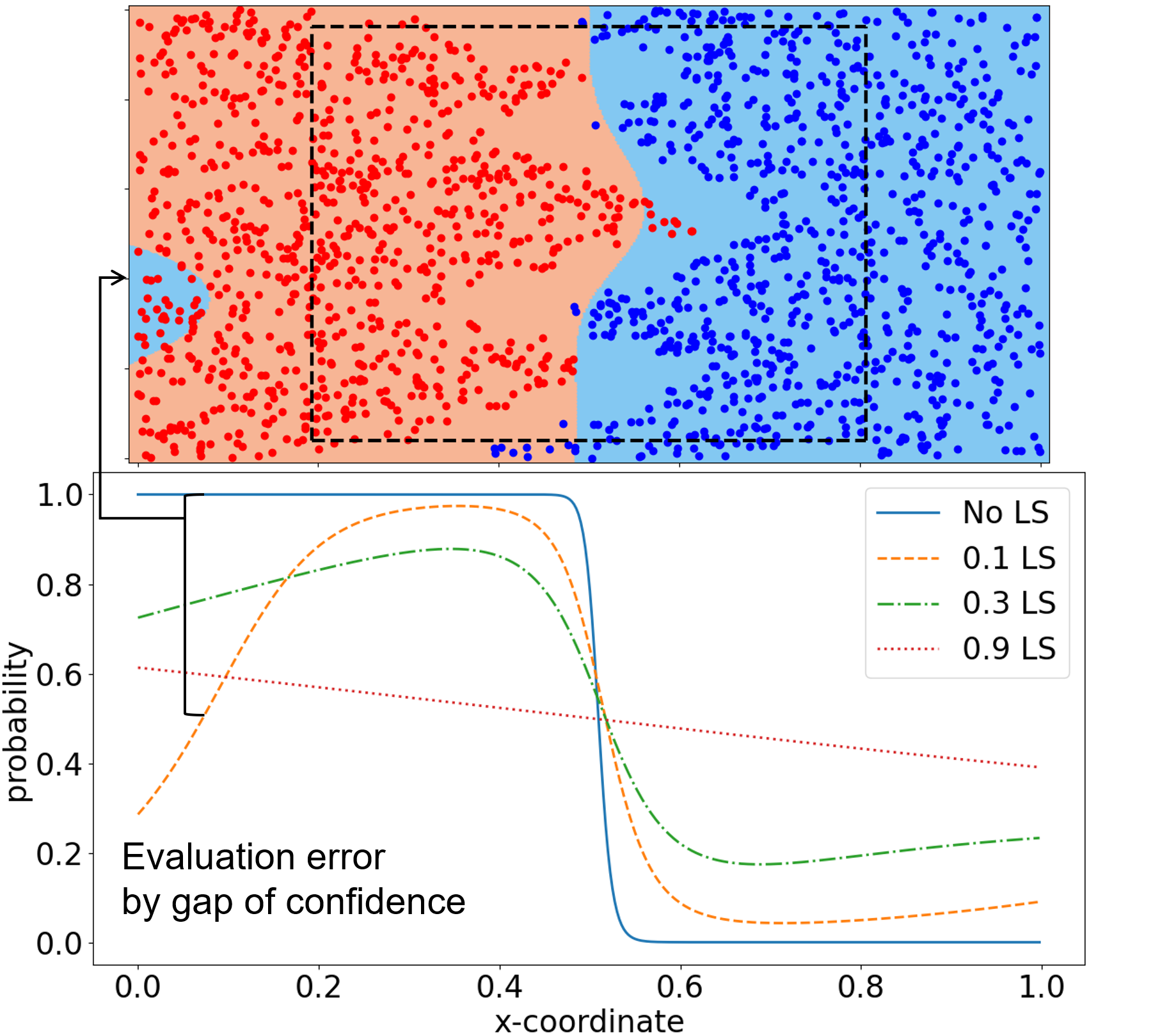}}
\caption{Confidence change of LS over seen and unseen areas in the sample space with respect to the interpolate rate. $n$ LS: LS of interpolate rate $n$ with uniform distribution.}
\label{LS_gap}
\end{figure}

LS can be interpreted as the interpolation of confidence scores with a uniform probability distribution over classes with generated label noise.
This approach reduces over-confidence in training data and induces a generalization effect. 
When learning the decision boundary, increasing the contribution of the uniform distribution smoothens the decision boundary by reducing the gap of confidence scores between samples differently classified. 
This effect helps find the simplest hypothesis, but it also causes a side effect of more sensitively when constructing decision boundaries to capture the regions of fluctuating loss.
If there is a small region on the sample area whose class is different from the majority of classes of the surrounding region, the difference over the two regions is amplified by strong LS. 
Thus, LS leads to a simpler decision boundary, but it is more sensitive to a small fluctuation of confidence as its strength increases.
Figure~\ref{LS_gap} shows an example case of our experiments for a toy problem, and the decrease of performance has been also reported in empirical works~\cite{chun2020empirical,muller2019does}.

\subsection{sKD-based Generalization}
With sKD-based generalization methods, teacher selection is the key factor to determining which hypotheses to train. 
Teachers in recently proposed methods can be split into four types: similar, pre-matured, random, and heuristic teachers. 
In Fig.~\ref{fig:penalization_geometric}(b), the example of the difference between teachers in a loss landscape is shown. 

\subsubsection{Similar Teacher}
A \textit{similar} teacher is a sufficiently trained model with similar generalization strength as the student model.
This teacher may be a model trained with a few prior updates before (i.e., Self-KD)~\cite{kim2020self} or a separately trained optimal model with a different random seed as Born-again-neural-network (BAN)~\cite{furlanello2018born}.
This teacher is naturally expected to improve performance because its effect is like the ensemble approach, which enhances robustness to seed variance. 
However, this benefit is a well-known post-hoc effect generally obtainable in most circumstances of applying neural networks and does not provide specific information about how to generalize a model in a plateau.

\subsubsection{Random Teacher}
LS is a type of sKD that interpolates confidence of prediction with a uniform distribution if we regard the uniform distribution of the confidence generated by the teacher. 
More directly, sKD can use a randomly initialized teacher without training (RI-KD). 
This teacher inherits the instability of LS because it randomly weakens the confidence distribution of a student model. 
It is further unstable because it allows many different models as a teacher whose gradients can largely vary over the plateau.
Moreover, in the loss landscape of Fig.~\ref{fig:penalization_geometric} (b), a random teacher can show a biased direction as penalization because of the small range of parameter values in the practical initialization of neural networks. 

\subsubsection{Pre-Matured Teacher}
\textit{Poorly trained} teachers have been used with sKD as a \textit{teacher-free} (TF)-KD$_{self}$~\cite{yuan2020revisiting}. 
The teacher is trained within only a few epochs so that it has some information about the training data, but its quality is low.
A random teacher allows many models and adds instability to guiding the student model; however, it can transfer some trained information.



\subsubsection{Heuristic Teacher}
In the same paper~\cite{yuan2020revisiting},
sKD using a heuristically designed pseudo teacher was also proposed, noted as TF-KD$_{reg}$. 
The teacher in the method has a specific probability of
\begin{equation}
\label{eq:tf_kd_reg}
p^d(k) = 
\begin{cases}
a &\text{if k = c} \\
\frac{1-a}{N-1} &\text{otherwise},
\end{cases},
\end{equation}
where $N$ is the number of classes, and $c$ is the correct label.
The generated probability is softened by Equation (\ref{eq:ssoftmax}) with $T$ as temperature.
This method results in a larger probability of a correct label, but the others have equal probability as LS.
Thus, it inherits the problem of random teachers.


\begin{figure}[htb]
\center{\includegraphics[width=0.5\textwidth]{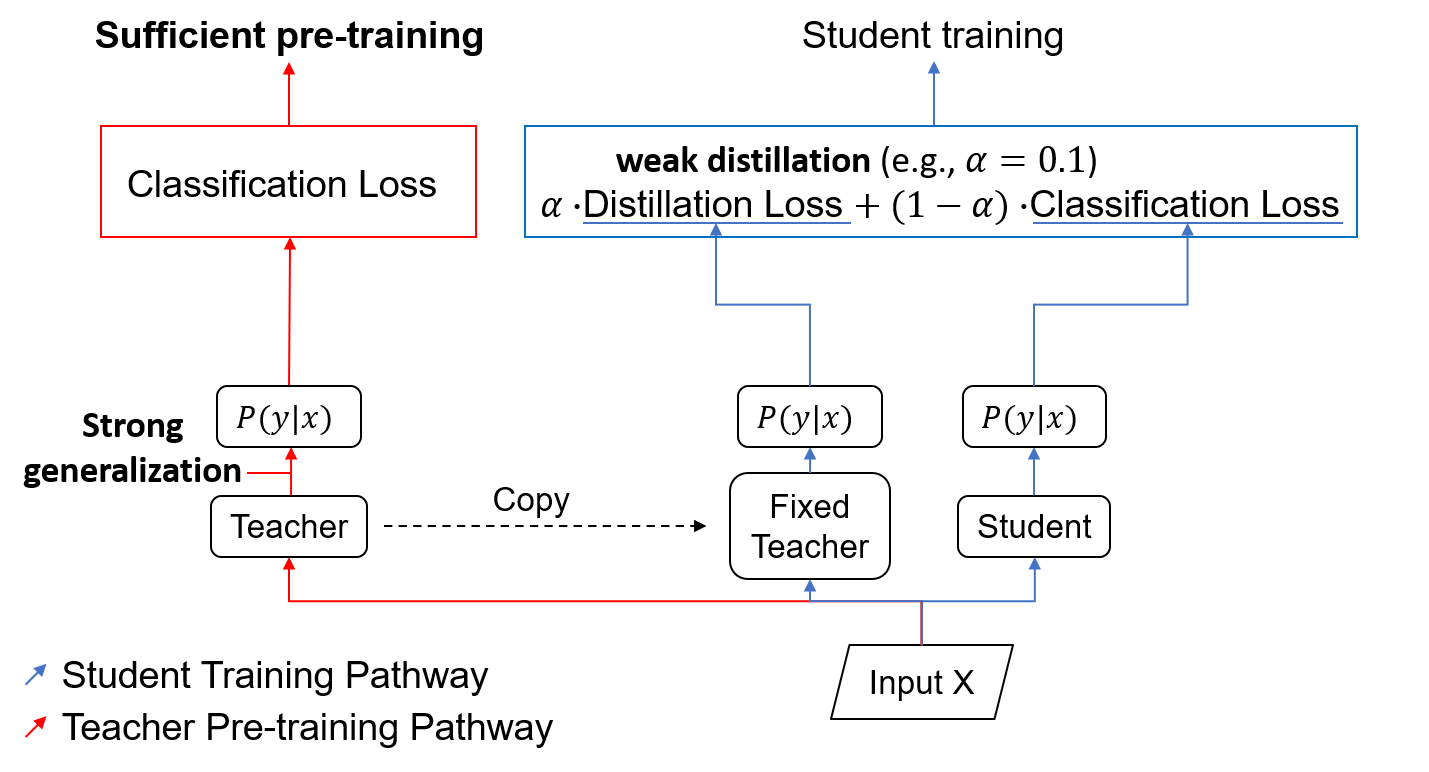}}
\caption{Process of the proposed method for generalization (y: ground truth; x: input sample; $\alpha$: hyperparameter of KD to imply transferring strength; Teacher: a neural network in the same architecture to Student)}
\label{fig:overall}
\end{figure}

\section{Method} \label{sec:method}
In this section, we propose a generalization method to use mDT-KD and its adaptive approach. 

\subsection{sKD Setting}
The existing definitions of sKD have been reported slightly differently by papers; hence we redefine it as a transference from an equally structured model after the teacher training is finished.
To transfer the intact dark knowledge, we use the equation (\ref{eq:naive_kd}) without the temperature term as
\begin{equation}
\label{eq:skd}
\mathcal{L}_{sKD} = (1 - \alpha) \cdot \mathcal{L}_{CE}(p_s, y_{gt}) + \alpha \cdot \mathcal{L}_{CE}(p_s, p_t),
\end{equation}
where $p_t$ and $p_s$ are probability vectors by teacher and student, and $\alpha$ is a hyperparameter used to control the interpolate rate.



\subsection{mDT-KD}
The whole process of mDT-KD is shown in Fig.~\ref{fig:overall}.
The distinguishing features of this method include the following conditions:
\begin{itemize}
\item pre-training: training a teacher until it reaches convergence with extremely strong generalization; and
\item distillation: transferring implicit knowledge on representations conservatively in sKD (extremely weak distillation).
\end{itemize}
The first condition is introduced to obtain a locally optimal model as simple as possible. 
To obtain such a model, we apply extremely strong generalization, which is rarely used in practical settings, and we train the model until it roughly converges to a certain loss. 
As shown in Fig.~\ref{fig:penalization_geometric} (b), mDT is likely to be located on an optimum in the area of more generalized models. 
The quality of the optimum is very low, but mDT is nonetheless a converged model that can provide more data-driven and less uncertain hypotheses than the other teachers.

The second condition guarantees that the teacher only affects the generalization on the uncertain region of decision boundaries. 
Because the teacher will be very inaccurate and may easily destroy the trained information of the student, the scale of distillation should be sufficiently small. 

To obtain a mDT, we recommend using dropout rather than penalization. 
Compared with the bias of updating decision boundaries of penalization methods, blurring confidence distribution near the boundaries in dropout  can be easily reduced by the weak distillation.

\subsection{Extension: Adaptive mDT-KD}
The mDT-KD may result in a small sacrifice in training accuracy because the mDT builds a confidence distribution critically bad for microscopic prediction. 
To overcome this problem, we propose an adaptive mDT-KD ($\textrm{mDT-KD}_{ada}$), which reduces the $\alpha$ gradually during training as the following equation:
\begin{equation}
\label{eq:ada_a}
\mathcal{L}_{sKD} = (1 - \rho\alpha) \cdot \mathcal{L}_{CE}(p_s, y_{gt}) + \rho\alpha \cdot \mathcal{L}_{CE}(p_s, p_t)
\end{equation}
where $\rho$ is a decaying constant equal to $\frac{\textrm{remaining epochs}}{\textrm{total epochs}}$.


\section{Experiments} \label{sec:exp}
We first analyzed decision boundaries and confidence distribution on a toy binary classification problem. 
To investigate the impact of generalization, we then evaluated test performance in three practical image classification tasks using MNIST ~\cite{lecun-mnisthandwrittendigit-2010}, CIFAR-10, and CIFAR-100~\cite{krizhevsky2009learning} datasets.
Details of the implemented codes and experiments are described in the supplementary material.

\subsection{Decision Boundary and Confidence Distribution Analysis}
\subsubsection{Toy Binary Classification Problem}
\begin{figure*}[!ht]
\center{\includegraphics[width=1\textwidth]{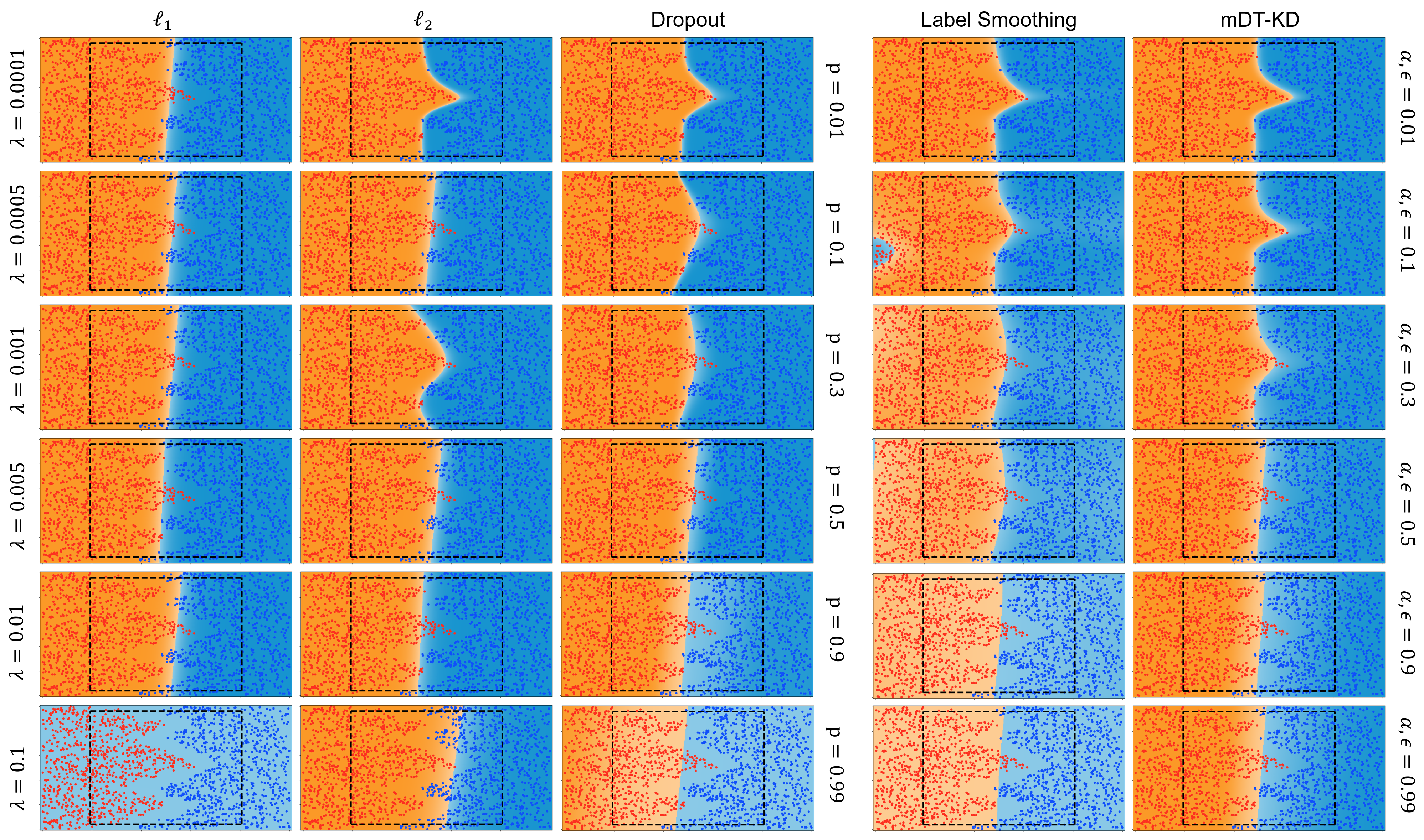}}
\caption{Decision boundary and confidence distribution of simple generalization (left three columns) and sKD-based generalization (right two columns) in the toy binary classification problem according to generalization strength. The red and blue dots are samples, and the background is a set of predictions of the location. Training samples are inside the black square and test samples are outside ($\lambda$: penalty scale; $p$:dropout probability; $\alpha$: interpolation rate of sKD in Equation~(\ref{eq:skd}); $\epsilon$: interpolation rate of LS in Equation~(\ref{eq:ls})). }
\label{fig:toy_result}
\end{figure*}
To analyze clear decision boundaries, we set a toy binary classification problem.
We first determined the correct nonlinear decision boundary to learn, and we then evenly generated samples to maintain the boundary. 
We exclusively distributed the samples for each class to the left or right area in 2-dimensional space so that we could find the proper simplest hypothesis to determine the decision boundary. 
Training samples were $1,000$ xy-coordinates labeled as zero or one. 
In the larger box, $760$ unseen samples were provided, which held the simplest decision boundary for the training data. 

\subsubsection{Model Setting}
To observe and analyze the clear effects of the generalization methods, we used a feed-forward neural network~(FFNN) using two layers; the hidden layer had five nodes. The activation function on the layer was the sigmoid function.
The output layer uses the softmax function. 

\subsubsection{Decision Boundary Evaluation}
Because a decision boundary comprises hyperplanes defined over all layers, we evaluated it using an inductive method. 
After training the network, we generated several grid points, and the network predicted their labels for detecting its trained decision boundary.

\subsubsection{Generalization Methods}
To understand the impact of common and primitive generalization methods on the trained decision boundary campared with the proposed method, we selected the $\ell_1$, $\ell_2$ penalization, dropout, LS, and mDT-KD. 
The generalization strength of the methods was set from the weakest to the strongest value to sufficiently break the decision boundary and confidence distribution.

\subsection{Performance in Practical Image Classification Tasks}
\subsubsection{Model and Data for MNIST}
As a pilot test, we trained FFNN with a hand-written MNIST dataset.
The model comprised three fully connected layers with $(28\times 28, 500)$, $(500, 500)$ and $(500, 10)$.
A rectified linear unit is used as the activation function, and $\ell_1$, $\ell_2$ and dropout were used for experiments.
Batch normalization~\cite{ioffe2015batch} was also used on each layer except not the output layer.
The model was optimized using stochastic gradient descent with a $0.01$ learning rate without momentum.
We used $60,000$ images for training and $10,000$ to test without a validation set.

\subsubsection{Model and Data for CIFAR-10 and CIFAR-100}
For generality, we also conducted experiments using convolutional neural networks on the image classification task.
We adopted WideResNet~(WRN) with and without our proposed method.
~The other training optimization and hyperparameters including $\ell_1$, $\ell_2$ and dropout, were the same with \cite{zagoruyko2016wide} using WRN-28-10, having 28 depth and 10 width.
We used $50,000$ images for training and $10,000$ to test without a validation set.

\begin{table*}[!ht]
\centering
	\caption{Top-one and -three test errors (\%) in image classification tasks. Statistics are evaluated from five runs with random seeds for each case. Self-KD~\cite{kim2020self}, TF-KD$_{self}$, and TF-KD$_{reg}$~\cite{yuan2020revisiting} are reproduced from the reported settings (Baseline: the best combination of three simple generalization methods; LS$^*$: the best model on the Baseline).}
\label{tab:performance_table_summary}
\begin{tabular}{c|l||rr|rr|rr}
&	\multicolumn{1}{c||}{Generalization}& \multicolumn{2}{c|}{MNIST} & \multicolumn{2}{c|}{CIFAR10} & \multicolumn{2}{c}{CIFAR100}\\
&\multicolumn{1}{c||}{method} &\multicolumn{1}{c}{top-1} &\multicolumn{1}{c|}{top-3} &\multicolumn{1}{c}{top-1} &\multicolumn{1}{c|}{top-3} &\multicolumn{1}{c}{top-1} &\multicolumn{1}{c}{top-3}\\
\hline
\hline
\multirow{3}{*}{\rotatebox[origin=c]{90}{\parbox[c]{12mm}{\centering Simple}}}
& Dropout  & 1.46 ($\pm 0.04$) & 0.13 ($\pm 0.02$) & 5.67 ($\pm 0.19$) & 0.68 ($\pm 0.08$)  & 23.17 ($\pm 0.08$) & 9.75 ($\pm 0.22$)  \\
& $\ell_1$ & 1.28 ($\pm 0.04$) & 0.13 ($\pm 0.03$) & 5.63 ($\pm 0.10$) & 0.68 ($\pm 0.08$)  & 22.45 ($\pm 0.16$) & 8.91 ($\pm 0.14$)  \\
& $\ell_2$ & 1.31 ($\pm 0.05$) & 0.17 ($\pm 0.01$) & 4.05 ($\pm 0.14$) & 0.50 ($\pm 0.05$)  & 19.13 ($\pm 0.09$) & 7.42 ($\pm 0.20$)   \\
& Baseline & 1.20 ($\pm 0.02$) & 0.13 ($\pm 0.02$) & 3.84 ($\pm 0.10$) & 0.42 ($\pm 0.06$)  & 18.70 ($\pm 0.13$) & 7.40 ($\pm 0.20$)  \\
\hline
\multirow{7}{*}{\rotatebox[origin=c]{90}{\parbox[c]{1cm}{\centering sKD}}}
& LS             & 1.40 ($\pm 0.06$)  & 0.18 ($\pm 0.04$)          & 5.61 ($\pm 0.13$)  & 0.91 ($\pm 0.04$)  & 21.98 ($\pm 0.26$) & 9.55 ($\pm 0.23$) \\
& LS$^*$         & \textbf{1.16} ($\pm 0.04$)  & 0.15 ($\pm 0.03$)          & 3.81 ($\pm 0.07$)  & 0.69 ($\pm 0.08$)  & 18.59 ($\pm 0.05$) & 7.54 ($\pm 0.21$)  \\
& BAN            & 1.23 ($\pm 0.03$)  & 0.12 ($\pm 0.02$)          & 3.84 ($\pm 0.07$)  & 0.43 ($\pm 0.03$)  & 18.62 ($\pm 0.13$) & 7.18  ($\pm 0.10$)  \\
& RI-KD          & 1.32 ($\pm 0.05$)  & 0.15 ($\pm 0.04$)          & 3.84 ($\pm 0.05$)  & 1.53 ($\pm 0.22$)  & 19.00 ($\pm 0.40$) & 10.61 ($\pm 0.62$) \\
& Self-KD        & 1.23 ($\pm 0.05$)  & 0.14 ($\pm 0.03$)          & 4.88 ($\pm 0.16$)  & 0.49 ($\pm 0.04$)  & 20.01 ($\pm 0.25$) & 7.36 ($\pm 0.23$) \\
& TF-KD$_{self}$ & 1.28 ($\pm 0.03$)  & \textbf{0.09} ($\pm 0.01$)          & 3.99 ($\pm 0.06$)  & 0.44 ($\pm 0.05$)  & 19.34 ($\pm 0.28$) & 7.60 ($\pm 0.13$) \\
& TF-KD$_{reg}$  & 1.47 ($\pm 0.09$)  & 0.18 ($\pm 0.05$)          & 3.97 ($\pm 0.04$)  & 0.75 ($\pm 0.07$)  & 19.69 ($\pm 0.29$) & 8.26 ($\pm 0.20$)  \\
\hline
\multirow{2}{*}{\rotatebox[origin=c]{90}{\parbox[c]{5mm}{\centering Ours}}}
& mDT-KD            & 1.19 ($\pm 0.02$) &  0.12 ($\pm 0.01$) & 3.75 ($\pm 0.06$)   & \textbf{0.40} ($\pm 0.05$)    & 18.46 ($\pm 0.19$) & 7.56 ($\pm 0.09$)     \\
& mDT-KD$_{ada}$    & 1.19 ($\pm 0.05$) &  0.15 ($\pm 0.03$) & \textbf{3.70} ($\pm 0.11$)  & 0.42 ($\pm 0.05$)    & \textbf{18.43} ($\pm 0.19$) & \textbf{7.13} ($\pm 0.15$)      \\
\hline
\end{tabular}
\end{table*}

\subsubsection{Generalization Methods}
We selected simple generalization and sKD-based generalization methods. 
The common methods included $\ell_1$, $\ell_2$ penalization, dropout, and their combination. 
The sKD-based methods are LS, $\textrm{LS}^*$, BAN, RI-KD, Self-KD, $\textrm{TF-KD}_{self}$, $\textrm{TF-KD}_{reg}$, mDT-KD, and $\textrm{mDT-KD}_{ada}$. 
The sKD-based models were trained with combinations of simple generalization methods as the baseline.
For implementation or re-production of all methods, details are shown in Table~\ref{tab:grid_search_range}.

\begin{table}[htb]

\centering
	\caption{Hyperparameter range for grid search of each generalization method. Intervals are in linear or logscale (LS$^*$: LS with the best combination of $\ell_1$, $\ell_2$ and dropout; h.p.: hyper-parameter).}
\label{tab:grid_search_range}

\begin{tabular}{llrc}
generalization & h.p. & range & \# of points\\
\hline
\hline
Dropout       & $p$           & $[0, 0.9]$ & 12\\
$\ell_1$      & $\lambda$     & $[10^{-7}, 5 \times 10^{-4}]$ & 8 \\
$\ell_2$      & $\lambda$     & $[10^{-6}, 5 \times 10^{-3}]$ & 8 \\
\hline
LS            & $\epsilon$    & $[10^{-4}, 0.5]$   & 10  \\
LS$^*$        & $\epsilon$    & $[10^{-4}, 0.5]$   & 10  \\
RI-KD         & $\alpha$      & $[10^{-2}, 0.3]$    & 4 \\
mDT-KD        & $p_t$         & $[0.5, 0.9999]$  & 8 \\
              & $\alpha$      & $[10^{-6}, 0.5]$   & 16\\
mDT-KD        & $p_t$         & $[0.5, 0.9999]$  & 8 \\
              & $\alpha$      & $[10^{-6}, 0.5]$   & 16\\
\end{tabular}

\end{table}

\begin{table}[htb]
\centering
\caption{Best hyperparameter settings found by the grid search for the test performance}
\label{tab:exp_hyperparam}
\begin{tabular}{l|l|rrr}
               && MNIST & CIFAR10  & CIFAR100  \\
\hline\hline
Dropout        & $p$                 & $3\mathrm{e}{-1}$ & $4\mathrm{e}{-1}$ & $1\mathrm{e}{-1}$    \\
$\ell_1$       & $\lambda_{\ell_1}$  & $5\mathrm{e}{-5}$ & $5\mathrm{e}{-6}$ & $5\mathrm{e}{-6}$    \\ 
$\ell_2$       & $\lambda_{\ell_2}$  & $1\mathrm{e}{-3}$ & $1\mathrm{e}{-3}$ & $1\mathrm{e}{-3}$    \\ 
Baseline       & $p$                 & $1\mathrm{e}{-2}$ & $3\mathrm{e}{-1}$ & $3\mathrm{e}{-1}$    \\ 
               & $\lambda_{\ell_1}$  & $1\mathrm{e}{-5}$ & 0                 & 0                    \\ 
               & $\lambda_{\ell_2}$  & $5\mathrm{e}{-4}$ & $5\mathrm{e}{-4}$ & $5\mathrm{e}{-4}$    \\ 
\hline
LS             & $\epsilon$          & $2\mathrm{e}{-1}$ & $1\mathrm{e}{-1}$ & $1\mathrm{e}{-1}$    \\
LS$^*$         & $\epsilon$          & $1\mathrm{e}{-3}$ & $1\mathrm{e}{-1}$ & $1\mathrm{e}{-1}$    \\
BAN            & $\alpha$            & $5\mathrm{e}{-1}$ & $5\mathrm{e}{-1}$ & $5\mathrm{e}{-1}$    \\
RI-KD          & $\alpha$            & $3\mathrm{e}{-1}$ & $1\mathrm{e}{-1}$ & $1\mathrm{e}{-1}$    \\
Self-KD        & $\alpha$            & $7\mathrm{e}{-1}$ & $7\mathrm{e}{-1}$ & $7\mathrm{e}{-1}$    \\
TF-KD$_{self}$ & $\alpha$            & $1\mathrm{e}{-1}$ & $1\mathrm{e}{-1}$ & $1\mathrm{e}{-1}$    \\
TF-KD$_{reg}$  & $\alpha$            & $1\mathrm{e}{-1}$ & $1\mathrm{e}{-1}$ & $1\mathrm{e}{-1}$    \\
               &          $\tau$     & 20                & 20                & 20                   \\
               &                 $a$ & 0.99              & 0.99              & 0.99                 \\
\hline
mDT-KD         & $p_t$               & 0.99              & 0.9               & 0.999                \\
               &       $\alpha$      & $5\mathrm{e}{-6}$ & $1\mathrm{e}{-2}$ & $1\mathrm{e}{-1}$    \\
mDT-KD$_{ada}$ & $p_t$               & 0.99              & 0.9               & 0.999                \\
               &       $\alpha$      & $1\mathrm{e}{-4}$ & $1\mathrm{e}{-1}$ &$5\mathrm{e}{-2}$
\end{tabular}
\end{table}

\subsubsection{Grid Search for Best Performance}
To compare the best performance of all methods, we applied grid search for the hyperparameters related to generalization, as shown in Table~\ref{tab:grid_search_range}. 
In the case of mDT-KD, we prepared mDTs sufficiently trained with various dropout probabilities. 
Then, the dropout probability of a teacher as involved for the grid search to find the best student. 
BAN~\cite{furlanello2018born}, Self-KD~\cite{kim2020self}, TF-KD methods~\cite{yuan2020revisiting} used the introduced settings from their respective papers. 
We ran each case once and all runs took about 30 days with 40 Geforce 2080TI GPUs.


\section{Results and Discussion} 
\label{sec:result}
\subsection{Decision Boundary and Confidence Distribution Analysis}
The decision boundary and confidence distribution results are illustrated in Fig.~\ref{fig:toy_result}.
The vertically lower setting has a stronger generalization, and the bottom line has nearly the largest strength.
The boundary of the orange and blue background represents the trained decision boundary. 
The lightness of the background colors is confidence in the sample. (e.g., the lighter area has lower confidence.)
The gap between boundary points of red and blue colors is the uncertain area for updating the decision boundary. 
The black dashed line is the boundary of inside training samples and outside unseen samples. 

In the results, we observed the simplest decision boundary in the problem, which penetrates the uncertain area. 
In the area, the penalization methods showed a larger change of the angle and the position of the decision boundary compared with dropout, LS, and mDT-KD, whereas the generalization strength increased. 
Dropout gradually and stably flattened the boundary with an increase of the generalization scale. 
The distinguished point was the rapid lightening of the background colors, which implied the confidence loss.
LS also showed stable decision boundary building in terms of generalization control, but the lighting appeared in the overall area in all conditions. 
Strangely, the second row image made a small false prediction of red points for unseen data. 
mDT-KD maintained a less flexible decision boundary than did the penalization methods. 
Dropout and LS also hold this property, but mDT-KD stably maintains overall confidence distribution. 

\subsection{Performance in Practical Image Classification Tasks}
\subsubsection{Test Performance}
In Table~\ref{tab:performance_table_summary}, the best accuracy results of each method in the grid search are shown with the applied hyperparameter values. 
mDT-KD methods showed the best accuracy in top one and top three accuracy results among all common and sKD-based generalization methods for CIFAR-10 and CIFAR-100 tasks. 
In the MNIST task, it showed the second best results lower than LS* with a difference of 0.03.
Compared with the baseline of the best $\ell_1$, $\ell_2$, and dropout combinations in the grid search, mDT-KD showed consistent improvements in all tasks.
$\textrm{mDT-KD}_{ada}$ with a decreasing $\alpha$ performed slightly better than did mDT-KD and was the best method. 
The other sKD-based methods were roughly better than the common methods, but their superiority was inconsistent. 
In the case of LS, the gap between LS and LS* was significantly large.

\subsubsection{Relation to Generalization Level of mDT}

\begin{figure*} 
\centering
\subfloat[MNIST\label{fig:dt_select_3d_a}]{%
 \includegraphics[width=0.3\textwidth]{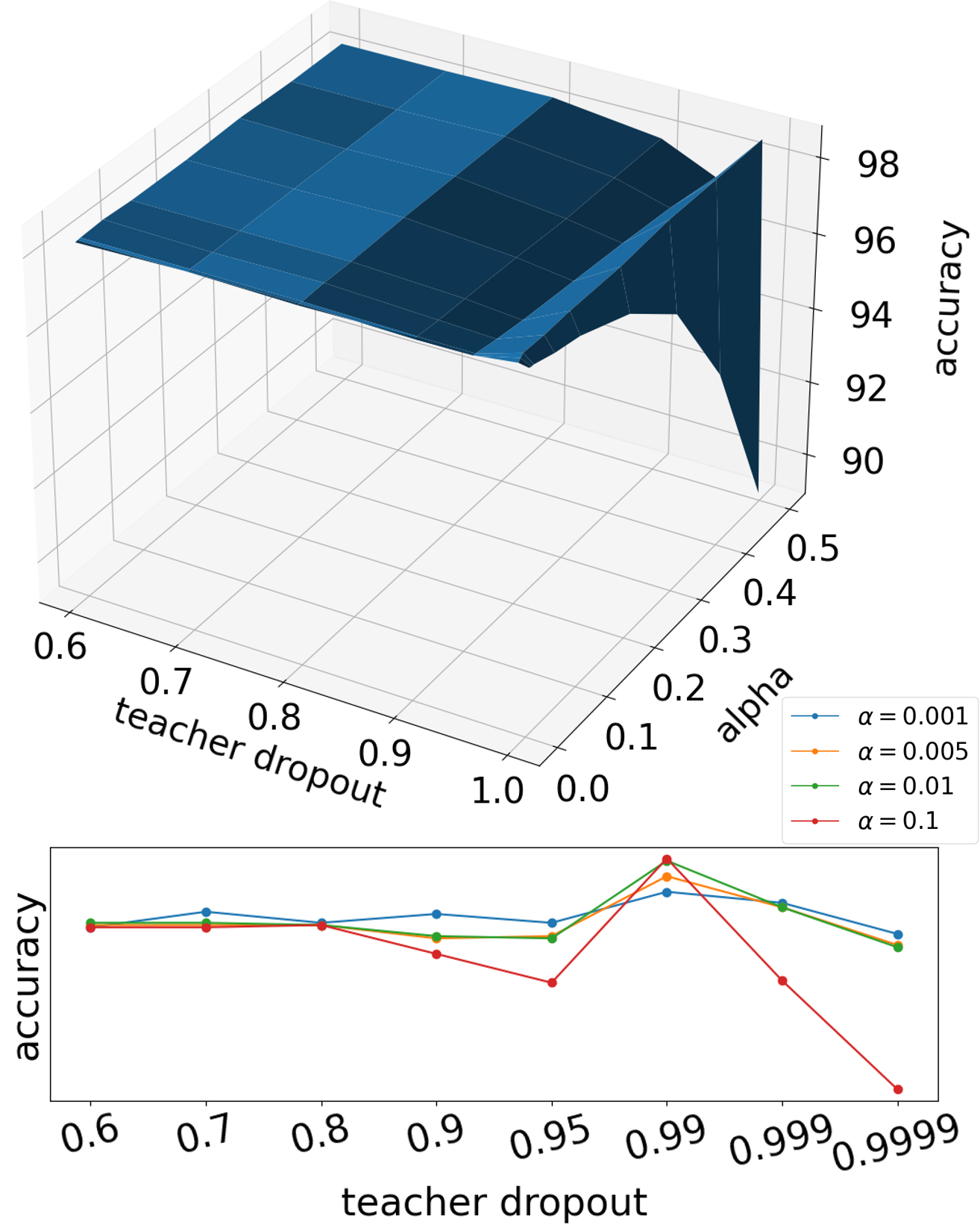}}
 \hfill
\subfloat[CIFAR-10\label{fig:dt_select_3d_b}]{%
\includegraphics[width=0.3\textwidth]{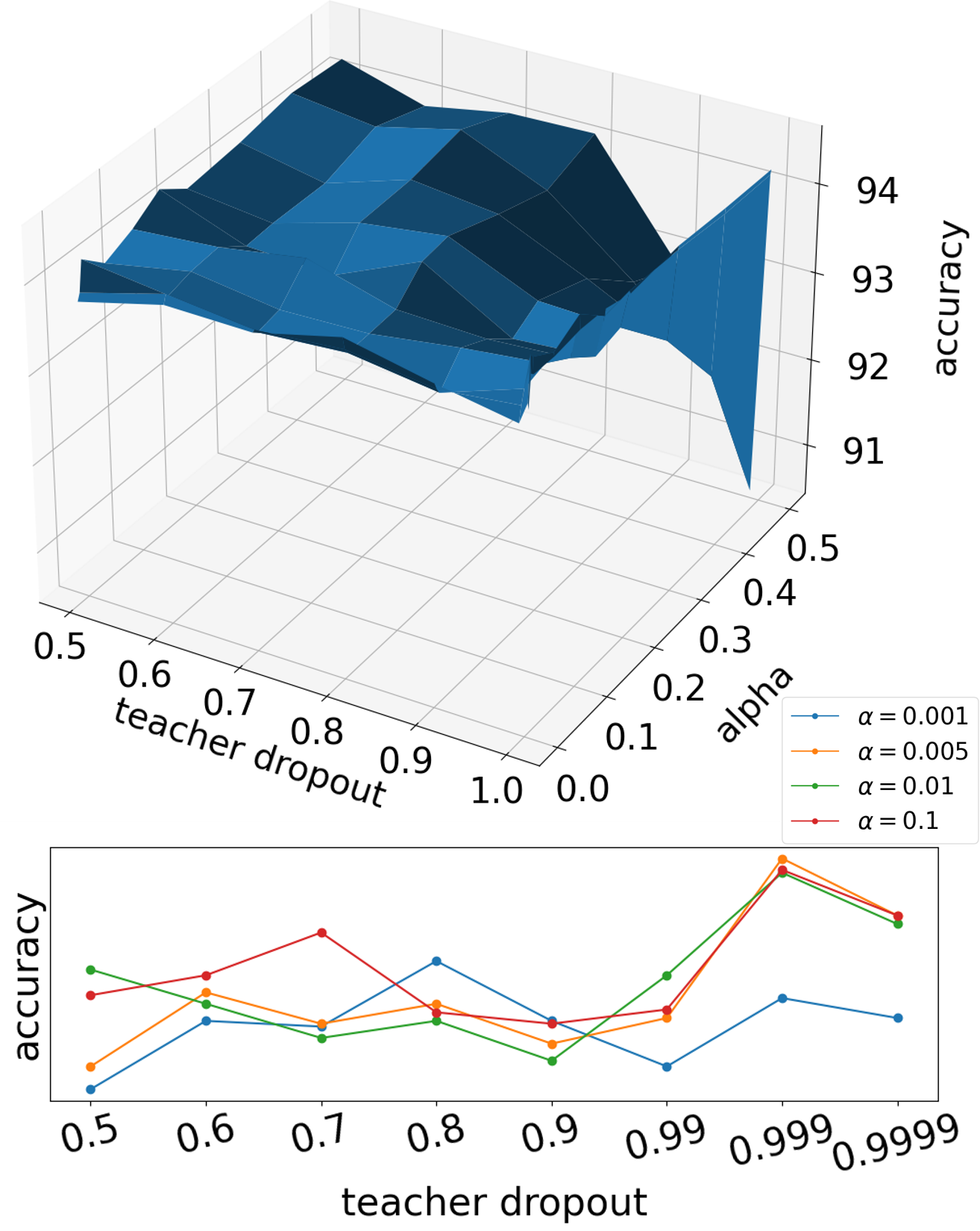}}
\hfill
\subfloat[CIFAR-100\label{fig:dt_select_3d_c}]{%
\includegraphics[width=0.3\textwidth]{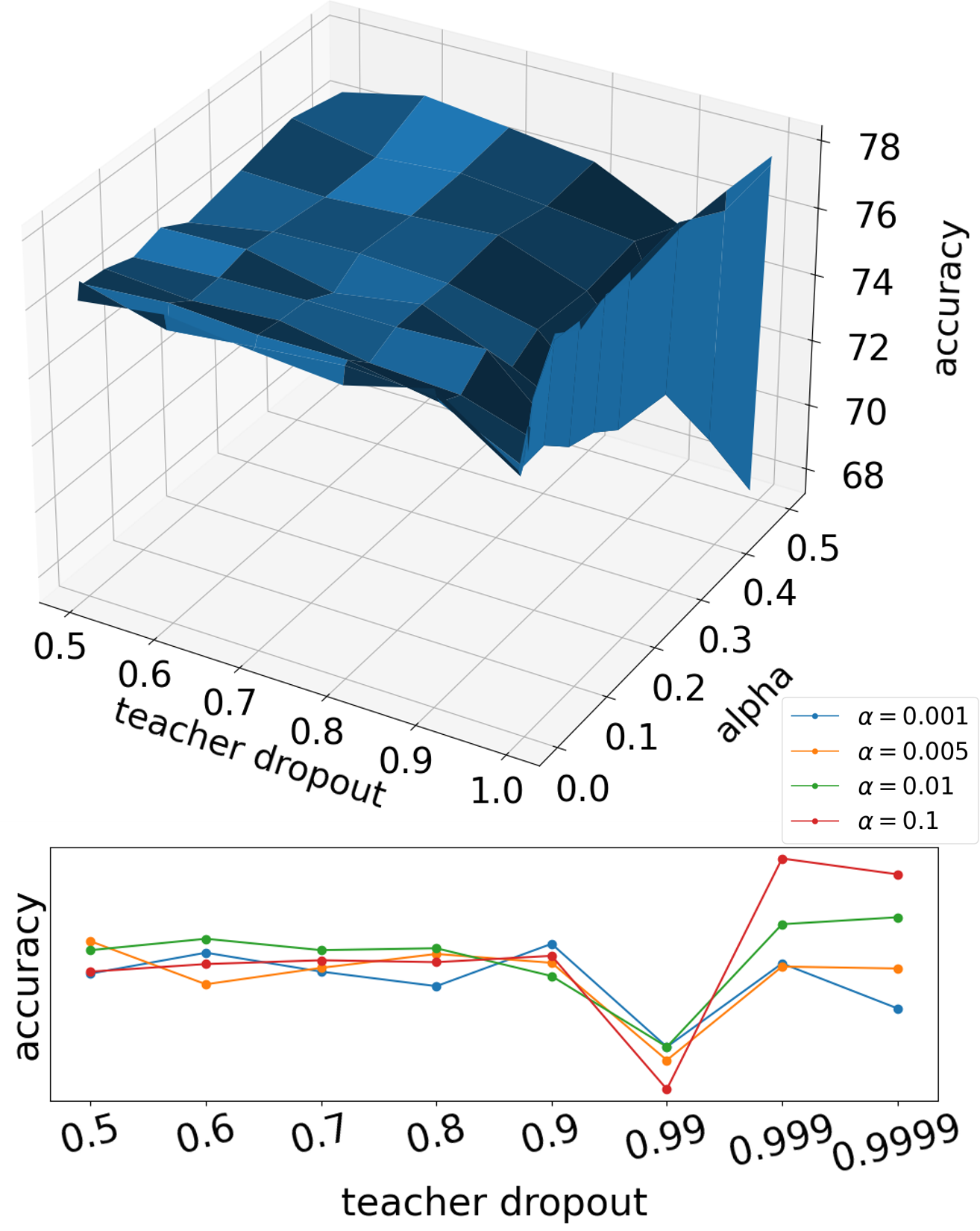}}
\caption{Sensitivity of mDT-KD to distillation strength and generalization level of teachers (alpha: dropout probability to turn off a node; accuracy: test accuracy of the student).} 
\label{fig:dt_select_3d} 
\end{figure*}
\begin{figure*} 
\centering
\subfloat[MNIST\label{fig:rbs_2d_a}]{%
 \includegraphics[width=0.3\textwidth]{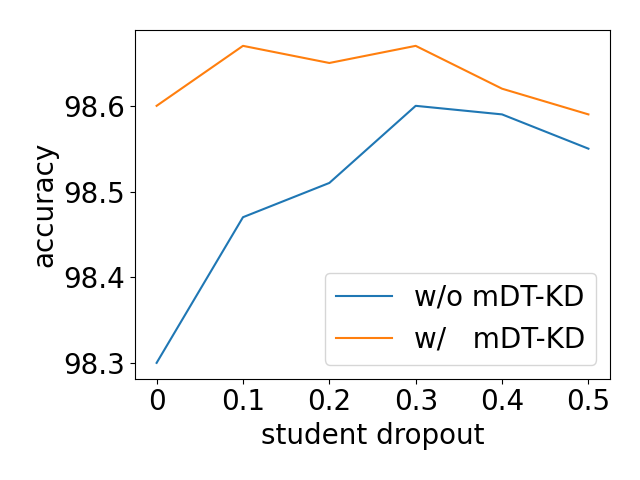}}
 \hfill
\subfloat[CIFAR-10\label{fig:rbs_2d_b}]{%
\includegraphics[width=0.3\textwidth]{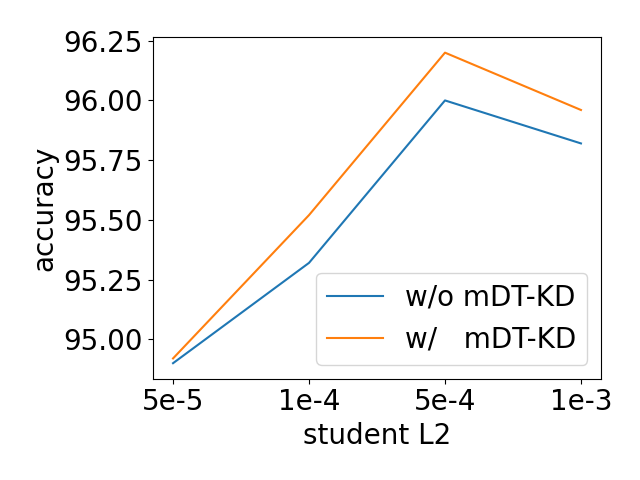}}
\hfill
\subfloat[CIFAR-100\label{fig:rbs_2d_c}]{%
\includegraphics[width=0.3\textwidth]{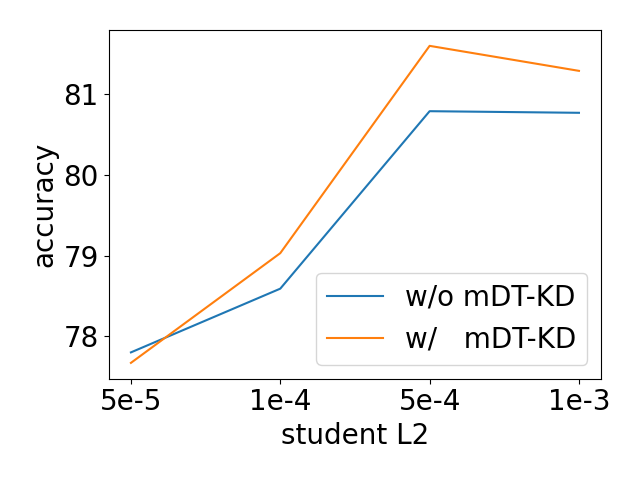}}
\caption{Robustness of mDT-KD to generalization level of base student models (the best simple generalization method and its hyperparameter for each task are selected in the grid search results). }
\label{fig:rbs_2d} 
\end{figure*}

Figure~\ref{fig:dt_select_3d} shows the best test accuracy with respect to the interpolate rate $\alpha$ and dropout probability of the matured dumb teacher model for mDT-KD. 
The mDTs of low dropout probability were excluded from the plots because of their largely lower accuracy. 
In the results, the best mDT was found when dropout probability is as extremely strong (i.e., 0.5 to 0.999) in all tasks. 
A valley exists between 0.9 and 1.0, and the best results are always located near the 1.0; however, the closest level was 0.9999, almost 1.0, which critically decreases accuracy.
The extremely generalized mDT consistently decreased the accuracy with an increase of $\alpha$ and the decrease speed. 
The supplementary material shows the numerical results of Figure~\ref{fig:dt_select_3d}. 

\subsubsection{Robustness to Generalization Level of the Base Student}
We evaluated the gain of the test accuracy by adding the mDT-KD method to already well-regularized student models. 
During the grid search for each task, the best student with the simple generalization was dropout for MNIST and $\ell_2$ for CIFAR-10 and CIFAR-100.
The test accuracy results of the baseline students and the best performance after adding mDT-KD are shown in Figure~\ref{fig:rbs_2d}. 
Although the generalization of students was stronger, the accuracy gradually decreased and was extremely low at the end. 
In the overall student settings, the best performance results were increased in all the tasks. 
When the generalization strength is weak, the performance stably increased. 
In the strong generalization condition, the performance was consistently lower than the baseline.

\section{Discussion} \label{sec:discussion}
In the toy task, we observed that mDT-KD had benefits by keeping the stable decision boundary and preserving the confidence distribution of students compared with the other generalization methods.

During the performance test, mDT-KD methods showed consistently better accuracy than the other simple generalization methods, even their best combination on all three tasks.
This observation demonstrates the effectiveness of the generalization, because training accuracy is almost zero in all cases. 
The slightly lower performance than LS* in MNIST may be explained by the lower possibility to have the uncertain region when the training data are sufficient to the simple task. 
The large variance of sKD-based methods raised the sensitivity to teacher selection. 
As reported, the random and poorly trained teachers improved the accuracy of some cases, but mDT can still increase performance. 
Furthermore, the mDT-KD methods showed their effectiveness, even with a sufficiently tuned student model as Baseline with simple generalization methods. 
The improvement in the above two environments implies that the uncertain region exists so that it can still be generalized by the simplest hypothesis. 
LS* is the most comparative method to mDT-KD, but it showed lower accuracy as the problem became more complex from CIFAR-10 to CIFAR-100.

Regarding the accuracy result related to the teacher generalization level, strong generalization showed roughly good performance, but the mDT near the 1.0 threshold was the best model. The almost random teacher performed closer to 1.0 than mDT, but showed far worse performance.
This result is consistent with other reports of random teacher performance improvement.
More importantly, it shows that there exists a better generalization guide built by data-driven mDT. 

In the robustness test to student generalization level, mDT-KD was shown to be consistently beneficial to already tuned models by a single and simple method with a large range of generalization strength. 
\section{Conclusion} \label{sec:conclusion}
In this paper, we investigated the effective assumptions related to generalizing an region of uncertainty wherein decision boundaries are not guided by training data. 
As a candidate assumption, we proposed \textit{the simplest hypothesis}, which is trained with strong generalization and sufficient convergence.
We clarified the limitations of both simple and sKD-based existing generalization methods in using the hypothesis.
To resolve the limitations, we proposed the \textit{matured Dumb Teacher based Knowledge Distillation} to conservatively apply the simplest hypothesis to a sKD framework. 
From the analysis of decision boundaries and confidence distributions, we have confirmed that mDT-KD stably adopts the simplest boundary without destroying the confidence distribution or extant data.
In more practical tests on MNIST, CIFAR-10, and CIFAR-100 datasets, mDT-KD consistently showed better generalization performance than simple and the sKD-based methods that operate on already sufficiently tuned students. 
In the deeper sensitivity analysis of teacher's generalization level, the existent of mDT induced the best generalization guide. 
These analysis and results imply that mDT-KD is effective to finely generalize a model while not destructing other trained and generalized information. 
This work can be extended to understand generalization effects in advanced KD tasks to directly transfer representations. 

\section*{Acknowledgement}
This work was supported by Institute of Information \& communications Technology Planning \& Evaluation (IITP) grant funded by the Korea government (MSIT) (R7119-16-1001, Core technology development of the real-time simultaneous speech translation based on knowledge enhancement), the Institute of Information \& communications Technology Planning \& Evaluation (IITP) grant funded by the Korea government (MSIT) (No. 2019-0-01842), Artificial Intelligence Graduate School Program (GIST), and the National Research Foundation of Korea (NRF) grant funded by the Korea government (MSIT) (2019R1A2C109107712).


%








\bibliographystyle{IEEEtran}
\bibliography{bib}
%




%

\begin{IEEEbiography}[{\includegraphics[width=1in,height=1.25in,clip,keepaspectratio]{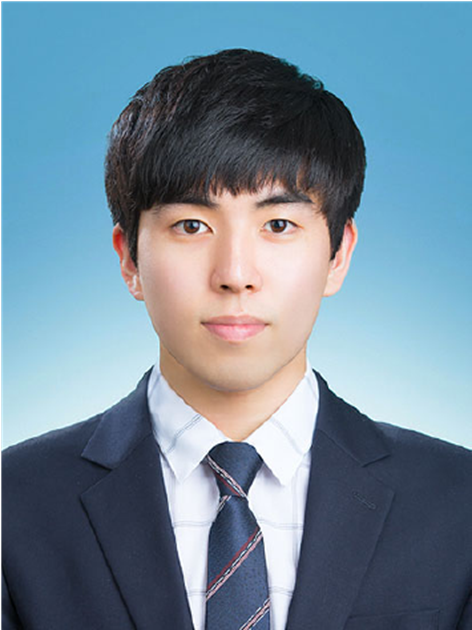}}]{HeeSeung Jung}
 is currently working toward the integrated Ph.D degree in Artificial Intelligence at Gwangju Institute of Science and Technology (GIST), Gwangju, Korea.
He received a B.Sc. degree in Department of Computer Science and Engineering from Konkuk University, Seoul, Korea, in 2020.
His research interests include artificial intelligence, machine learning, and natural language processing.
\end{IEEEbiography}

\begin{IEEEbiography}[{\includegraphics[width=1in,height=1.25in,clip,keepaspectratio]{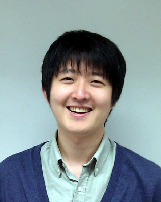}}]{Kangil Kim}
 received the B.Sc. degree in computer science from Korea Advanced Institute of Science and Technology, Daejeon, Korea, in 2006 and the Ph.D. degree from Seoul National University, Seoul, Korea, in 2012.
He was a Senior Researcher with the Natural Language Processing Group, Electronics and Telecommunications Research Institute, Seoul until 2016 and an assistant professor of Computer Science and Engineering Department at Konkuk University until 2019. 
He is currently an assistant professor of Electronics Engineering and Computer Science Department and Artificial Intelligence Graduate School in Gwangju Institute of Science and Technology.  
His research interests include artificial intelligence, evolutionary computation, machine learning, and natural language processing.
\end{IEEEbiography}


\begin{IEEEbiography}[{\includegraphics[width=1in,height=1.25in,clip,keepaspectratio]{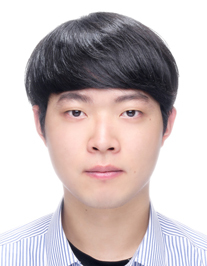}}]{Hoyong Kim}
 received the B.S degree in software from the Ajou University, Suwon, South Korean, in 2018, and the M.S degree from University of Science and Technology, Daejeon, South Korea, in 2020. He was a Research Student at Research Data Sharing Center, Korea Institute of Science and Technology Information, Daejeon, until 2020. He is currently the Ph.D student at Artificial Intelligence Graduate School, Gwangju Institute of Science and Technology, Gwangju, South Korea. His research interests include artificial intelligence, machine learning and natural language processing.
\end{IEEEbiography}

\begin{IEEEbiography}[{\includegraphics[width=1in,height=1.25in,clip,keepaspectratio]{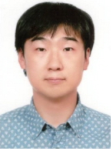}}]{Jong-hun Shin} received his B.Sc degree in computer and multimedia enginnering from the Pukyong National University, Busan, Rep. of Korea in 2008, and his MS degree in computer science from the Pusan National University, Busan, Rep. of Korea, in 2010. Since 2010, he has been working as a senior researcher with the Language-Intelligence Research Section, Electronics and Telecommunications Research Institute, Daejeon, Rep. Of Korea. His major research interests include artificial intelligence, machine translation, and natural-language processing.
\end{IEEEbiography}



\end{document}


\maketitle

In this supplementary material, we provide the following information to clarify our results: 
\begin{enumerate}
    \item details of the implemented codes and experiments;
    \item time and space complexity compared to common regularization methods;
    \item used hyper-parameters for grid-search in experiments to obtain the best performance of each regularization methods; and
    \item detailed numerical results of the grid search used for drawing Fig. 7. 
\end{enumerate}

\section{Experiments Details}
All of the experiments are implemented with Python 3.8 and PyTorch on Ubuntu 18.04.
An experiment for MNIST needs about 2 hours during $500$ epochs and for CIFAR needs about 8 hours during $200$ epochs.
There are about $1,500$ training models for each experiment, MNIST, CIFAR-10, and CIFAR-100.
We adjusted the random seed variance five times so that there are about $7,500$ models for MNIST and $150,000$ models for CIFAR.
40 Geforce 2080TI GPUs are used to train the models for about 30 days.

\section{Time and Space Complexity}

Our approach to use mDT-KD takes $\mathcal{O}(B)$ time-and-space complexity for model training because it must learn $\mathcal{O}(B)$ twice for training the dumb teacher and the student model from it.
When searching the best configuration of regularization, our method uses the same number of hyperparameters compared with the other KD-based methods and label smoothing in the condition of using extremely regularized dropout for mDT.
Thus, the grid-search cost is as practical as the other KD applications. 
\section{Hyperparameters for Grid Search}

\begin{table}[htb]

\centering
	\caption{Grid-search range for regularization methods. Linear and logscale intervals are used in the ranges. LS$^*$ is the best combination of LS and other common regularization methods. h.p. means hyperparameter.}
\label{tab:grid_search_range}

\begin{tabular}{llrc}
regularization & h.p. & set\\
\hline
\hline
Dropout       & $p$           & \{ 0, 0.01, 0.05, 0.1, 0.2, 0.3, 0.4, 0.5, 0.6, 0.7, 0.8, 0.9 \}\\
$\ell_1$      & $\lambda$     & \{ $10^{-7}$, $5 \cdot 10^{-7}$, $10^{-6}$, $5 \cdot 10^{-6}$, $10^{-5}$, $5 \cdot 10^{-5}$, $10^{-4}$, $5 \cdot 10^{-4}$ \}\\
$\ell_2$      & $\lambda$     & \{ $10^{-6}$, $5 \cdot 10^{-6}$, $10^{-5}$, $5 \cdot 10^{-5}$, $10^{-4}$, $5 \cdot 10^{-4}$, $10^{-3}$, $5 \cdot 10^{-3}$ \} \\
\hline
LS            & $\epsilon$    & \{ $0.0001$, $0.001$, $0.005$, $0.01$, $0.05$, $0.1$, $0.2$, $0.3$, $0.4$, $0.5$ \}  \\
LS$^*$        & $\epsilon$    & \{ $0.0001$, $0.001$, $0.005$, $0.01$, $0.05$, $0.1$, $0.2$, $0.3$, $0.4$, $0.5$ \}  \\
RI-KD         & $\alpha$      & \{ $0.01$, $0.1$, $0.2$, $0.3$ \}  \\
mDT-KD        & $p_t$         & \{ $0.5$, $0.6$, $0.7$, $0.8$, $0.9$, $0.99$, $0.999$, $0.9999$ \} \\
              & $\alpha$      & \{ $10^{-6}$, $5 \cdot 10^{-6}$, $10^{-5}$, $5 \cdot 10^{-5}$, $10^{-4}$, $5 \cdot 10^{-4}$, $10^{-3}$, $5 \cdot 10^{-3}$, $10^{-2}$, $5 \cdot 10^{-2}$, $0.1$, $0.15$, $0.2$, $0.3$, $0.4$, $0.5$ \} \\
mDT-KD$_{ada}$& $p_t$         & \{ $0.5$, $0.6$, $0.7$, $0.8$, $0.9$, $0.99$, $0.999$, $0.9999$ \} \\
              & $\alpha$      & \{ $10^{-6}$, $5 \cdot 10^{-6}$, $10^{-5}$, $5 \cdot 10^{-5}$, $10^{-4}$, $5 \cdot 10^{-4}$, $10^{-3}$, $5 \cdot 10^{-3}$, $10^{-2}$, $5 \cdot 10^{-2}$, $0.1$, $0.15$, $0.2$, $0.3$, $0.4$, $0.5$ \} \\
\end{tabular}

\end{table}

\section{Numerical Results of Grid Search Shown in Fig. 7}
\begin{table*}[h]
\centering
\begin{tabular}{l|llllllll}
\diagbox[]{$\alpha$}{$p_t$} 
& 0.6    & 0.7    & 0.8    & 0.9    & 0.95   & 0.99   & 0.999  & 0.9999 \\
\hline
0.001 & 98.24 & 98.31 & 98.26 & 98.30 & 98.26 & 98.40 & 98.35 & 98.21  \\
0.005 & 98.25 & 98.25 & 98.25 & 98.19 & 98.20 & 98.47 & 98.33 & 98.16  \\
0.01  & 98.26 & 98.26 & 98.25 & 98.20 & 98.19 & 98.54 & 98.33 & 98.15  \\
0.05  & 98.29 & 98.24 & 98.27 & 98.15 & 98.09 & 98.52 & 98.15 & 97.95  \\
0.1   & 98.24 & 98.24 & 98.25 & 98.12 & 97.99 & 98.55 & 98.00 & 97.51  \\
0.2   & 98.23 & 98.24 & 98.28 & 98.05 & 97.85 & 98.53 & 97.42 & 96.28  \\
0.3   & 98.28 & 98.22 & 98.27 & 97.95 & 97.71 & \textbf{98.63} & 96.28 & 93.57  \\
0.4   & 98.23 & 98.22 & 98.26 & 97.94 & 97.54 & 98.59 & 93.54 & 87.11  \\
0.5   & 98.24 & 98.23 & 98.26 & 97.92 & 97.28 & 98.57 & 89.11 & 79.14  \\

\end{tabular}
\caption{Accuracy values of MNIST in Fig.~7, where $p_t$ is the dropout probability of a teacher, and $\alpha$ is the teacher rate in Equation~5.}
\end{table*}

\begin{table}[h]
\centering
\begin{tabular}{l|llllllll}
\diagbox[]{$\alpha$}{$p_t$} 
& 0.5    & 0.6    & 0.7    & 0.8    & 0.9    & 0.99   & 0.999  & 0.9999 \\
\hline
0.001 & 93.75 & 93.99 & 93.97 & 94.20 & 93.99 & 93.83 & 94.07 & 94.00  \\
0.005 & 93.83 & 94.09 & 93.98 & 94.05 & 93.91 & 94.00 & \textbf{94.56} & 94.36  \\
0.01  & 94.17 & 94.05 & 93.93 & 93.99 & 93.85 & 94.15 & 94.51 & 94.33  \\
0.05  & 93.94 & 94.12 & 93.98 & 94.00 & 94.13 & 94.22 & 94.36 & 94.32  \\
0.1   & 94.08 & 94.15 & 94.30 & 94.02 & 93.98 & 94.03 & 94.52 & 94.36  \\
0.15  & 94.29 & 93.93 & 93.88 & 94.30 & 93.74 & 93.80 & 94.26 & 94.34  \\
0.2   & 94.07 & 93.95 & 94.22 & 94.34 & 93.91 & 93.87 & 94.29 & 94.26  \\
0.3   & 94.14 & 93.94 & 94.24 & 94.12 & 93.68 & 93.24 & 94.26 & 94.31  \\
0.4   & 94.26 & 94.13 & 94.35 & 93.93 & 93.17 & 92.35 & 94.19 & 94.26  \\
0.5   & 94.22 & 93.93 & 94.10 & 94.12 & 92.87 & 90.51 & 94.17 & 94.23 
\end{tabular}
\caption{Accuracy values of CIFAR10 in Fig.~7, where $p_t$ is the dropout probability of a teacher, and $\alpha$ is the teacher rate in Equation~5.}
\end{table}

\begin{table}[h]
\centering
\begin{tabular}{l|llllllll}
\diagbox[]{$\alpha$}{$p_t$} 
& 0.5    & 0.6    & 0.7    & 0.8    & 0.9    & 0.99   & 0.999  & 0.9999 \\
\hline
0.001 & 76.09 & 76.42 & 76.12 & 75.89 & 76.56 & 74.93 & 76.25 & 75.54  \\
0.005 & 76.60 & 75.92 & 76.18 & 76.40 & 76.26 & 74.72 & 76.20 & 76.17  \\
0.01  & 76.46 & 76.64 & 76.46 & 76.49 & 76.05 & 74.93 & 76.87 & 76.98  \\
0.05  & 76.40 & 76.20 & 76.36 & 75.89 & 75.73 & 74.98 & 78.01 & \textbf{78.21}  \\
0.1   & 76.12 & 76.24 & 76.30 & 76.27 & 76.37 & 74.26 & 77.91 & 77.66  \\
0.15  & 76.37 & 76.44 & 75.89 & 76.29 & 75.22 & 74.00 & 77.95 & 78.03  \\
0.2   & 75.91 & 76.34 & 76.41 & 76.26 & 75.66 & 73.40 & 78.02 & 77.79  \\
0.3   & 76.26 & 76.15 & 76.29 & 76.43 & 75.51 & 73.11 & 78.20 & 78.03  \\
0.4   & 76.66 & 76.15 & 76.36 & 76.07 & 75.73 & 70.40 & 77.32 & 77.91  \\
0.5   & 76.30 & 76.85 & 76.55 & 76.20 & 74.97 & 67.45 & 77.46 & 77.73 
\end{tabular}
\caption{Accuracy values of CIFAR100 in Fig.~7, where $p_t$ is the dropout probability of a teacher, and $\alpha$ is the teacher rate in Equation~5.}
\end{table}